\documentclass{article}
\usepackage{spconf,amsmath,graphicx}

\usepackage{amsmath,graphicx, color, soul, multirow}
\usepackage{amssymb}
\usepackage{subfig}
\usepackage{bm}
\usepackage{verbatim}
\usepackage{cite}
\usepackage{amssymb}
\usepackage{pifont}
\usepackage{float}
\usepackage{amsmath}

\title{DIALOG-CONTEXT AWARE END-TO-END SPEECH RECOGNITION}
\name{Suyoun Kim$^1$ and Florian Metze$^2$}
\address{
  $^1$Electrical \& Computer Engineering\\
  $^2$Language Technologies Institute, School of Computer Science\\
  Carnegie Mellon University \\
  \texttt{\{suyoung1{\textbar}fmetze\}@andrew.cmu.edu}
}

\begin{document}
%
\maketitle
\begin{abstract}
Existing speech recognition systems are typically built at the sentence level, although it is known that dialog context, e.g. higher-level knowledge that spans across sentences or speakers, can help the processing of long conversations. The recent progress in end-to-end speech recognition systems promises to integrate all available information (e.g. acoustic, language resources) into a single model, which is then jointly optimized. It seems natural that such dialog context information should thus also be integrated into the end-to-end models to improve further recognition accuracy.  
In this work, we present a dialog-context aware speech recognition model, which explicitly uses context information beyond sentence-level information, in an end-to-end fashion.
Our dialog-context model captures a history of sentence-level contexts, so that the whole system can be trained with dialog-context information in an end-to-end manner. We evaluate our proposed approach on the Switchboard conversational speech corpus, and show that our system outperforms a comparable sentence-level end-to-end speech recognition system.
\end{abstract}
\section{INTRODUCTION}
\label{sec:intro}

\begin{figure*}[t]
  \centering
  \includegraphics[width=4in]{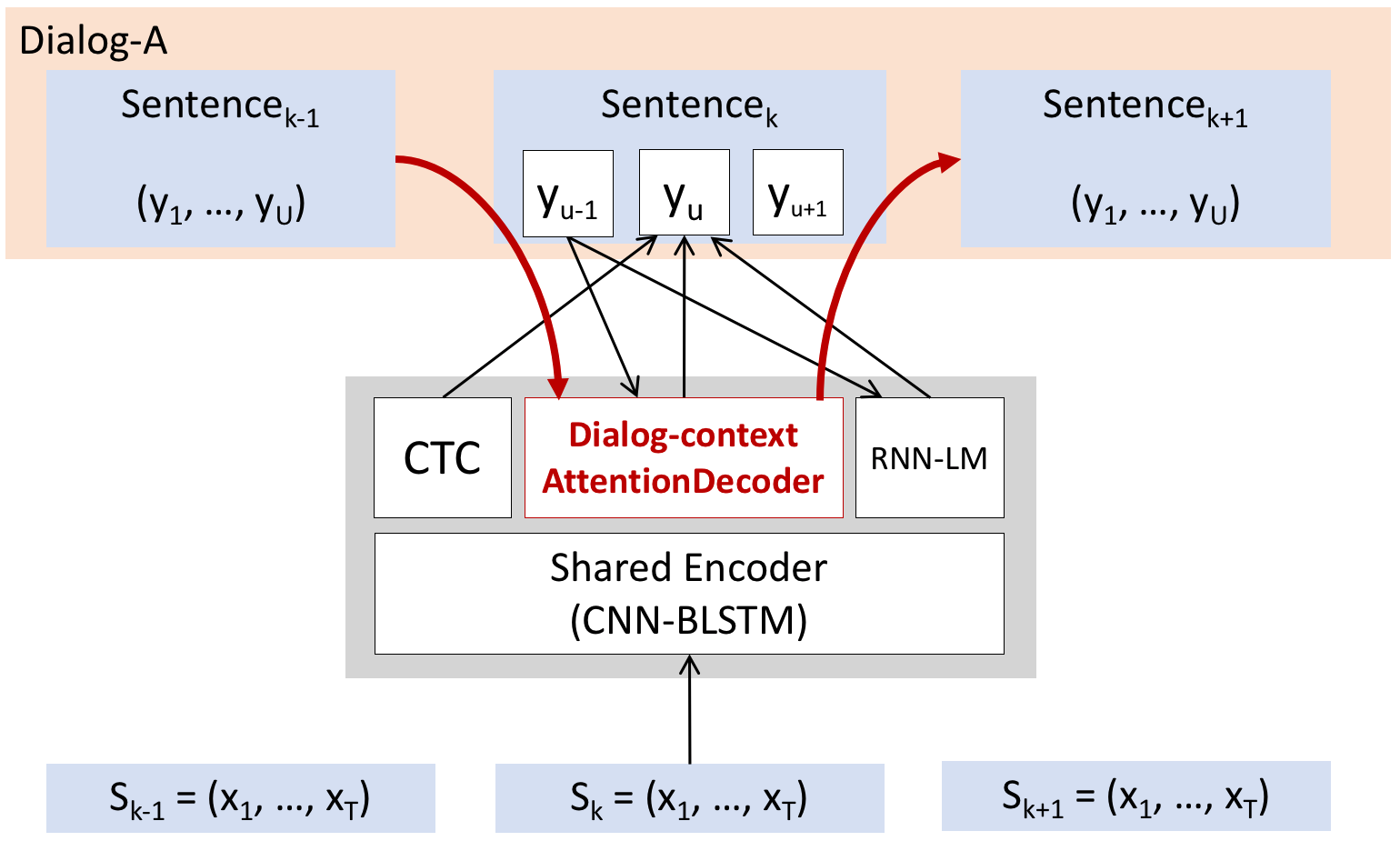}
  
\caption{The architecture of our dialog-context end-to-end speech recognition model. The red curved line represents the context information flow.}
\label{fig:arch}
\end{figure*}

As voice-driven interfaces to devices become mainstream, spoken dialog systems that can recognize and understand long dialogs are becoming increasingly important. Dialog context, dynamic contextual flow across multiple sentences,
provides important information that can improve speech recognition. 
To build speech recognition models, the long dialogs typically split into short utterance-level audio clips to make training the speech recognition models computationally feasible. Thus, such speech recognition models built on sentence-level speech data may lose important dialog-context information. 

In language model trained only on text data, several recent studies have attempted to use document-level or dialog-level context information to improve language model performance. Recurrent neural network (RNN) based language models \cite{mikolov2010recurrent} have shown success in outperforming conventional n-gram based models due to their ability to capture long-term information. Based on the success of RNN based language models, recent research has developed a variety of ways to incorporate document-level or dialog-level context information \cite{mikolov2012context, wang2015larger, ji2015document, liu2017dialog}. Mikolov et al. proposed a context-dependent RNN language model \cite{mikolov2012context} using a context vector which is produced by applying latent Dirichelt allocation \cite{blei2003latent} on the preceding text. Wang et al. proposed using a “bag-of-words” to represent the context vector \cite{wang2015larger}, and Ji et al. proposed using the last RNN hidden states from the previous sentence to represent the context vector \cite{ji2015document}. Liu et al. proposed using an external RNN to model dialog context between speakers \cite{liu2017dialog}. All of these models have been developed and optimized on text data, and therefore must still be combined with conventional acoustic models, which are optimized separately without any context information beyond sentence-level. The recent study \cite{xiong2017microsoft} attempted to integrated such dialog session-aware language model with acoustic models, however, it requires disjoint training procedure. 

There have been no studies of speech recognition models that incorporate dialog-context information in end-to-end training approach on both speech and text data. The recently proposed \textit{end-to-end} speech recognition models, a neural network is trained to convert a sequence of acoustic feature vectors into a sequence of graphemes (characters) rather than senones. Unlike sequences of senone predictions, which need to be decoded using a pronunciation lexicon and a language model, the grapheme sequences can be directly converted to word sequences without any additional models. The end-to-end models proposed in the literature to use a Connectionist Temporal Classification framework \cite{graves2006connectionist, graves2014towards, hannun2014deep, miao2015eesen, zweig2017advances}, an attention-based encoder-decoder framework \cite{bahdanau2014neural, chorowski2014end, chorowski2015attention, chan2015listen}, or both \cite{kim2017joint, watanabe2017hybrid}. 

Our goal is to build speech recognition model that explicitly use a dialog-level context information beyond sentence-level information especially in an end-to-end manner so that the whole system can be trained with the long context information. In this paper, we present a dialog-context aware end-to-end speech recognition model that can capture a history of sentence-level contexts within an end-to-end speech recognition models. We also present a method to \textit{serialize} datasets for training and decoding based on their onset times to learn dialog flow. We evaluate our proposed approach on the Switchboard conversational speech corpus \cite{swbd, godfrey1992switchboard}, and show that our model outperforms the sentence-level end-to-end speech recognition model.



\section{DIALOG-CONTEXT END-TO-END ASR}
\label{sec:review}


\subsection{End-to-end models}
We perform end-to-end speech recognition using a joint CTC/Attention-based approach with graphemes as the output symbols \cite{kim2017joint, watanabe2017hybrid}. The key advantage of the joint CTC/Attention framework is that it can address the weaknesses of the two main end-to-end models, Connectionist Temporal Classification (CTC) \cite{graves2006connectionist} and attention-based encoder-decoder (Attention) \cite{bahdanau2015end}, by combining the strengths of the two. With CTC, the neural network is trained according to a maximum-likelihood training criterion computed over all possible segmentations of the utterance's sequence of feature vectors to its sequence of labels while preserving left-right order between input and output. With attention-based encoder-decoder models, the decoder network can learn the language model jointly without relying on the conditional independent assumption.

Given a sequence of acoustic feature vectors, $\bm x$, and the corresponding graphemic label sequence, $\bm y$, the joint CTC/Attention objective is represented as follows by combining two objectives with a tunable parameter $\lambda: 0 \leq \lambda \leq 1$:
	\begin{align}
	    \label{e12}
		\mathcal{L} &= \lambda \mathcal{L}_\text{CTC} + (1-\lambda) \mathcal{L}_\text{Attention}.
	\end{align}
Each loss to be minimized is defined as the negative log likelihood of the ground truth character sequence $\bm{y^*}$, is computed from:
    \begin{align}
        \begin{split}
        \mathcal{L}_\text{CTC} \triangleq & -\ln \sum_{\bm{\pi} \in \Phi(\bm{y})} p(\bm{\pi}|\bm{x}) 
        \end{split}
    \end{align}
    \begin{align}
        \begin{split}
		\mathcal{L}_\text{Attention} \triangleq & -\sum_u \ln p(y_u^*|\bm{x},y^*_{1:u-1}) 
        \end{split}
    \end{align}
where $\bm{\pi}$ is the label sequence allowing the presence of the \textit{blank} symbol, $\Phi$ is the set of all possible $\bm{\pi}$ given $u$-length $\bm{y}$, and $y^*_{1:u-1}$ is all the previous labels.

\subsection{Dialog-context models}
\label{sec:dcrnnlm}
In this section, we present our \textit{DialogAttentionDecoder} subnetwork. We extend the \textit{AttentionDecoder} which is a subnetwork of standard end-to-end models in order to employ context information. The core modeling idea of this work is to integrate the context information from the previous sentence into that of the current sentence. 

Let we have a dataset consists of N-number of dialogs, $D = \{ d_1, \cdots, d_N \}$ and each dialog $d_i = (s_1, \cdots, s_K)$ has K utterances. $k$-th utterance $s_k$ is represented as a sequence of $U$-length output characters ($\bm{y}$) and $T$-length input acoustic features ($\bm{x}$). Given the high-level representation ($\bm{h}$) of input acoustic features ($\bm{x}$) generated from \textit{Encoder},
both the standard \textit{AttentionDecoder} and our proposed \textit{DialogAttentionDecoder} generates the probability distribution over characters ($ ~ y_u$), conditioned on ($\bm{h}$), and all the characters seen previously ($y_{1:u-1}$). Our proposed \textit{DialogAttentionDecoder} additionally conditioning on dialog context vector ($c_k$), which represents the information of the preceding utterance in the same dialog as:
    \begin{align}
	    \label{e6}
		\bm{h}       = & \textit{ Encoder}(\bm{x}) \\
	    \label{e7}
	    y_u \sim &
	        \begin{cases}
	            \text{standard decoder network:}\\
	            \qquad \textit{ AttentionDecoder}(\bm{h}, y_{1:u-1})\\
	            \text{proposed decoder network:}\\
	            \qquad \textit{ DialogAttentionDecoder}(\bm{h}, y_{1:u-1}, c_k)\\
	        \end{cases} \\
	\end{align}	

\begin{figure}[H]
\begin{minipage}[b]{1.0\linewidth}
  \centering
  \centerline{\includegraphics[width=8.5cm]{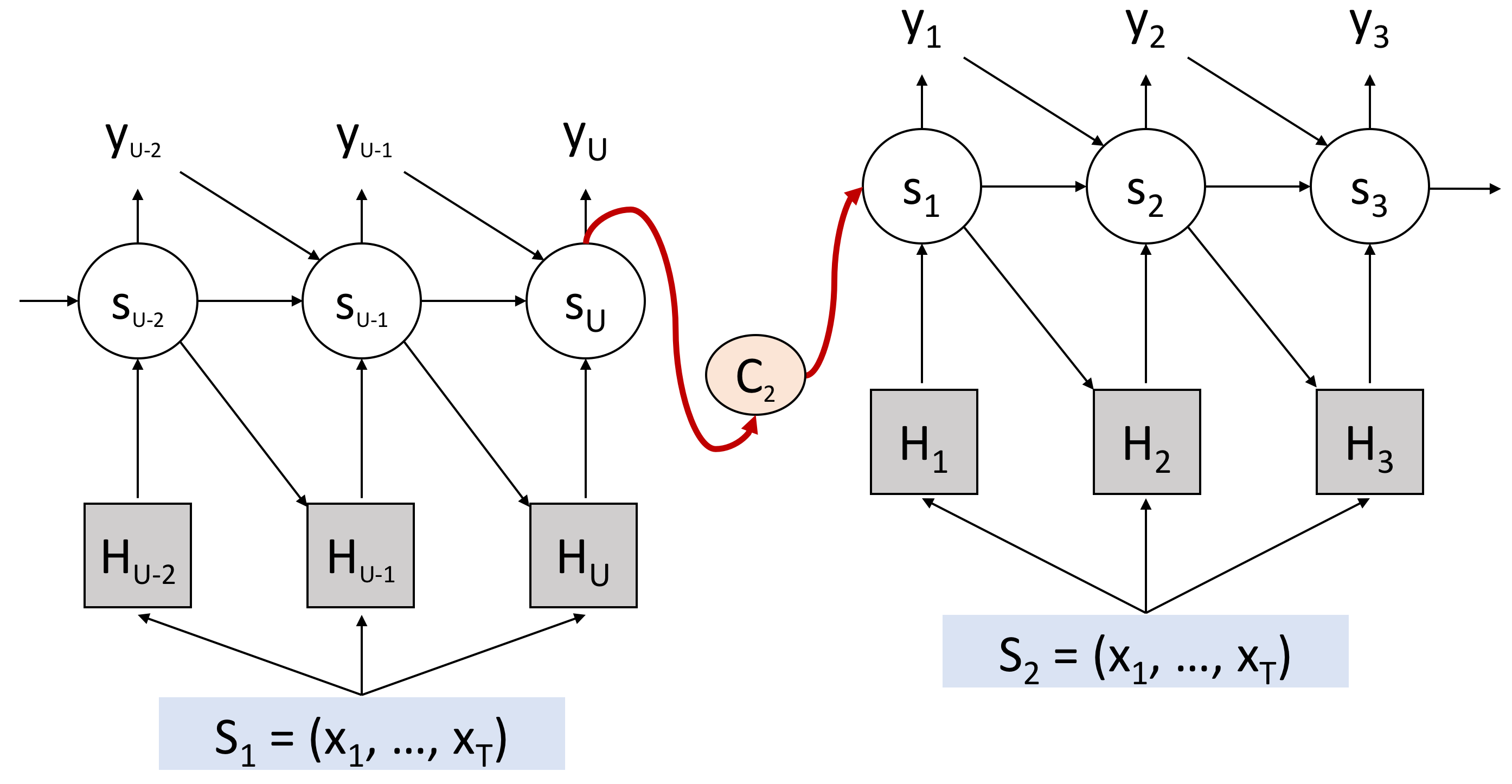}}
  (a) \\
  \centerline{\includegraphics[width=8.5cm]{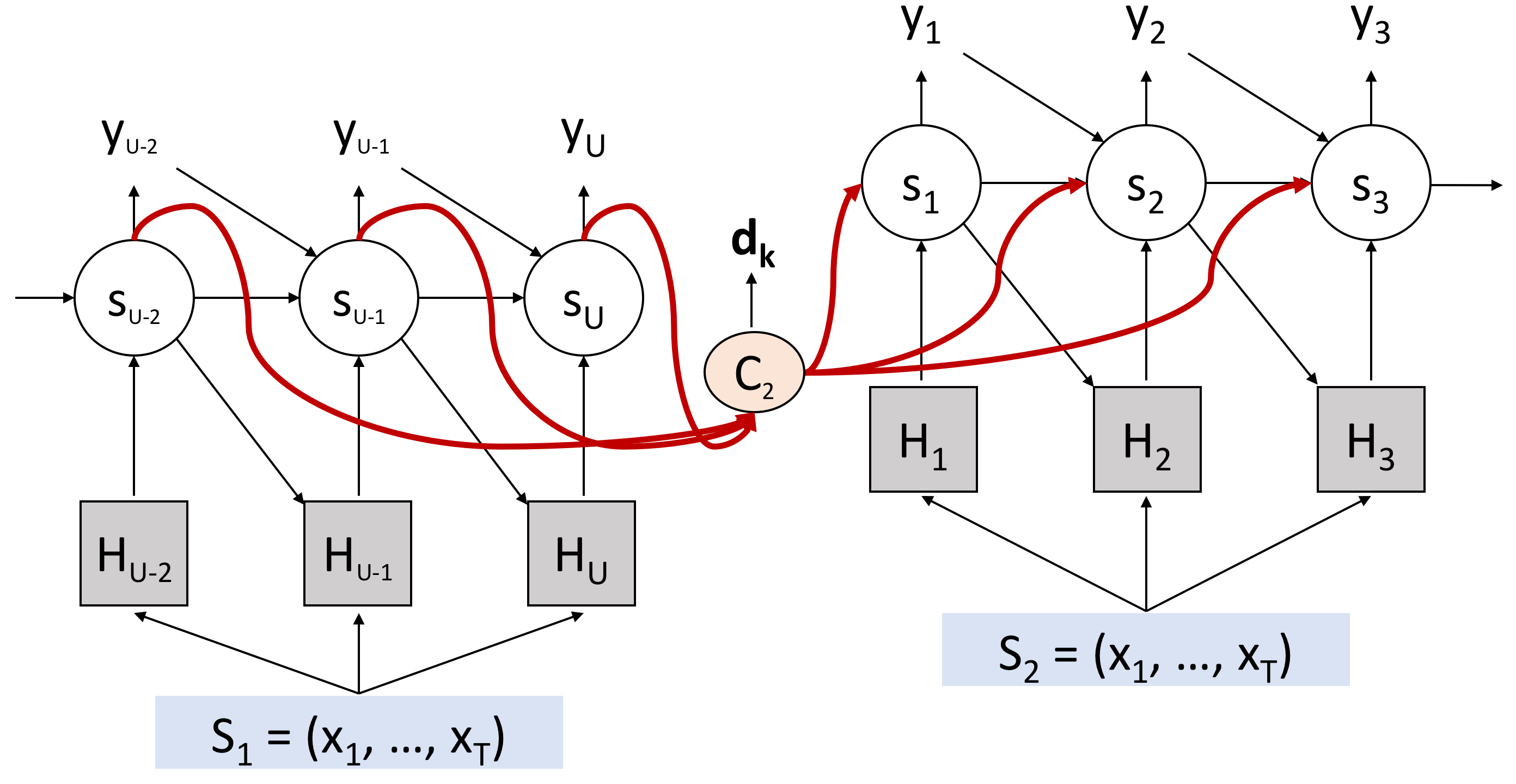}}
  (b) \\
\end{minipage}
\caption{Two different types of our dialog-context AttentionDecoder network. At each output step ($u$), the character distribution ($y_u$), is generated conditioning on 1) the dialog context vector ($c_{k}$) from the previous sentence, in addition to 2) the attended inputs ($H_u$) from Encoder network, 3) the decoder state ($s_{u-1}$), and 4) the history of character $y_{u-1}$. The red curved line represents the context information flow.}
\label{fig:dec}
\end{figure}

We represent the context vector, $c_k$, in two different ways as illustrated in Figure \ref{fig:dec}. In method (a), the last decoder state of the previous sentence represents the context vector, $c_k$. The context vector $c_k$ is propagated to the initial decoder state of current sentence. In method (b), every output information is integrated with additional attention mechanism and represents the context vector, $c_k$, then is propagated to every decoder state for each output time step of current sentence. 
Motivated by prior work \cite{ji2015document}, we produce the context vector for the (a)-type of our decoder from the final hidden representation of the previous sentence $s_{k-1}$:
	\begin{align}
		c_k &= s_U^{s_{k-1}}.
    \end{align}
For the (b)-type of our decoder, we embed a subnetwork (\textit{DialogContextGenerator}) with an additional attention mechanism to incorporate every previous context into a single context vector $c_k$. We use the character distributions of previous sentence ($y*_1 \cdots y*_U$) as input to generate ($c_k$). This subnetwork is optimized towards minimizing the dialog classification error:
    \begin{align}
        d_k \sim & \textit{DialogContextGenerator}(\bm{y*}) \\
        \mathcal{L}_\text{Dialog} \triangleq & - \ln p(d_k^*|\bm{y*})
    \end{align}
The additional loss $\mathcal{L}_\text{Dialog}$ is added to the $\mathcal{L}$ in Eq. (1) so that the whole model is optimized jointly. Once the context vector of previous sentence is generated by either proposed method (a) or (b), it is stored for the next sentence prediction. 


The following equation represents how to update the hidden states of \textit{DialogAttentionDecoder} with high-level input features ($\bm{h_u}$) generated from \textit{Encoder}, 2) the previous character ($y_{u-1}$), and 3) the context vector ($c_k$) from previous sentence:
	\begin{align}
	    \Hat{s}_{u-1} &=  f(s_{u-1}, c_k) \\
		s_u &= RNN( \Hat{s}_{u-1} , \bm{h_u}, y_{u-1}))  \\
		y_u  & \sim \text{softmax}(s_u)
    \end{align}
where $f(\cdot)$ is a function that combines the two inputs, $s_{u-1}$, and $c_k$:
\begin{align}
     f(s_{u-1}, c_k)  = \tanh (W s_{u-1} + V c_k + b)
\end{align}
where $W, V, b$ are trainable parameters. In this work, we use $\tanh$ for the non-linear activation function.

\subsection{Dataset serialization}
\label{sec:serial}
In order to learn and use the dialog context during training and decoding, we serialized the sentences based on their onset times and their dialogs rather than random shuffling of data. We first grouped the sentences based on their dialogs, then ordered the sentences according to their onset times. Instead of generating a minibatch set in a typical way that randomly chooses sentences, we created a minibatch to contain the sentences from each one of different dialogs. For example, a size-30 minibatch had 30 sentences from $d_1$ – $d_{30}$. We did not shuffle the minibatch sets for training or decoding, so that the context information generated from the previous minibatch that contains preceding sentences can propagate to the next minibatch. 

Since the number of sentences varies across the dialogs, some dialogs may not have enough sentences to construct the minibatches. In this case, we included dummy input/output data for the dialogs that have fewer sentences to maintain the minibatch size and not to lose context information of the other dialogs that have more sentences. We then masked out the loss from the dummy data for the objective function. Once every sentence in $d_1$ – $d_{30}$ was processed, then the sentences from the other dialogs $d_{31} - d_{60}$ would be processed. 

\begin{figure}[t]
\begin{minipage}[b]{1.0\linewidth}
  \centering
  \centerline{\includegraphics[width=8.5cm]{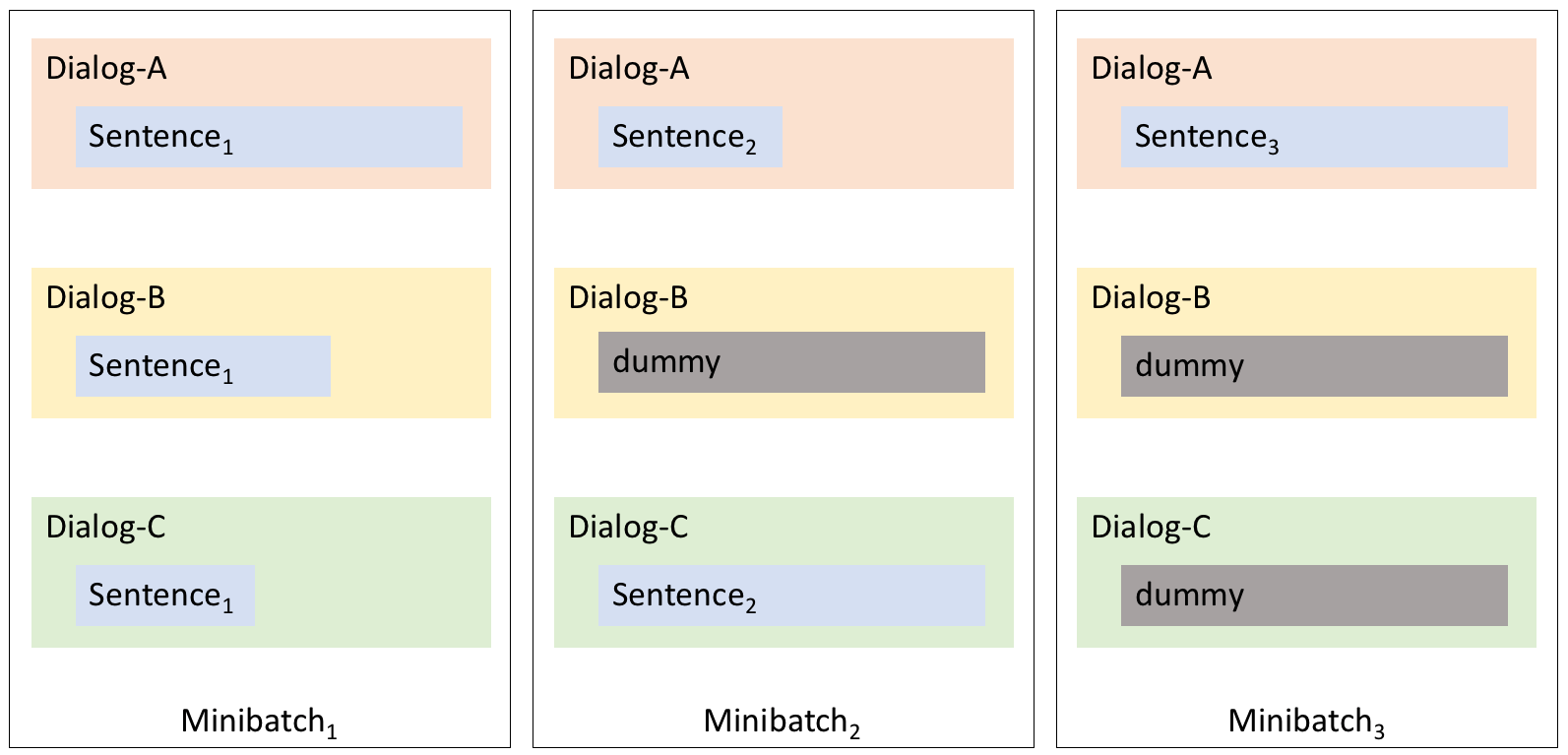}}
\end{minipage}
\caption{A method to make the minibatch set for training and evaluating our models. The example minibatches are serialized according to their onset times and their dialogs. The size of example minibatch is 3 and the sentences are from three dialogs, $d_A$, $d_B$, and $d_C$. In second and third minibatches, the dummy input/output data is included in the position for the $d_B$ and $d_C$, which have fewer sentences.}
\label{fig:serial}
\end{figure}

Figure \ref{fig:serial} illustrates the example minibatches that are serialized according to their onset times and their dialogs. The size of minibatch is 3 and the sentences are from dialogs, $d_A$, $d_B$, and $d_C$. In second and third minibatches, the dummy input/output data is included in the position for the $d_B$ and $d_C$, which have fewer sentences.

\section{EXPERIMENTS}
\label{sec:exp}

\subsection{Experimental corpora}

We investigated the performance of the proposed dialog-context aware model on the Switchboard LDC corpus (97S62) which has a 300 hours training set. Note that we did not use the Fisher dataset. We split the Switchboard data into two groups, then used 285 hours of data (192 thousand sentences) for model training and 5 hours of data (4 thousand sentences) for hyper-parameter tuning. Evaluation was carried out on the HUB5 Eval 2000LDC corpora (LDC2002S09, LDC2002T43), which have 3.8 hours of data (4.4 thousand sentences), and we show separate results for the Callhome English (CH) and Switchboard (SWB) evaluation sets. We denote train\_nodup, train\_dev, SWB, and CH as our training, development, and two evaluation datasets for CH and SWB, respectively. Table \ref{tab:dialog} shows the number of dialogs per each dataset.

\begin{table}[t]
\caption{ Experimental dataset description. We used the Switchboard dataset which has a 300 hours training set. Note that any pronunciation lexicon or external text data was not used.}
\label{tab:dialog}
\begin{center}
\begin{tabular}{|r|r|r|r|r|}
\hline
&Train\_nodup & Train\_dev & CH & SWB        \\
\hline
\hline
\# dialog     & 2,402      & 34       & 20   & 20 \\
\hline
\# sentences & 192,656    & 4,000    & 2,627 & 1,831 \\
\hline
\end{tabular}
\end{center}
\end{table}

We sampled all audio data at 16kHz, and extracted 80-dimensional log-mel filterbank coefficients with 3-dimensional pitch features, from 25~ms frames with a 10ms frame shift. We used 83-dimensional feature vectors to input to the network in total. We used 49 distinct labels: 26 characters, 10 digits, apostrophe, period, dash, underscore, slash, ampersand, noise, vocalized-noise, laughter, unknown, space, start-of-speech/end-of-speech, and blank tokens. 

Note that no pronunciation lexicon was used in any of the experiments. 

\begin{table*}[t]
\centering

\caption{Comparison of hypothesis between baseline and our proposed model. Three examples of two consecutive sentences are manually chosen from evaluation dataset. They show that our model correctly predicted the word bolded in current sentence, while the baseline incorrectly predicted it. The preceding sentence includes the context information related to the word, and our model seems to be benefit this information.}
\label{tab:example}
\resizebox{\textwidth}{!}{
\begin{tabular}{|r|l|l|}
\hline
model    & previous sentence   & current sentence                                          \\
\hline
\hline
REF      & \textit{well    claire's kindergarten but     she     is      already past    the                                       } & \textit{\textbf{kindergarten } i       mean                                 }\\
Baseline & \textit{well    clears in the garden     but     she     is      already past                                            } & \textit{\textbf{in a garden }     i       mean                             }\\
Ours     & \textit{well clears   kindergarten but     she     is      already past    the                                          } & \textit{\textbf{kindergarten} i       mean                                 }\\
\hline
REF      & \textit{yes    it      is      so      hot     in      the     building  have you ever been in it }                       & \textit{it      is      like    a       \textbf{sauna}}                     \\
Baseline & \textit{yeah if when he      said that   is      like    just   so      hot     in      the     belly have ever been but} & \textit{it is like i  \textbf{saw}                          }             \\
Ours     & \textit{yeah if when he is  in this  like    it      is      so      hot     in      the belief I have never been but }   & \textit{it      is      like    a      \textbf{ sauna    }   }              \\
\hline
REF      & \textit{if      we      go      we      like    check   into    a       to  a       to      a  hotel   but              } & \textit{i       know   but it is      much    more    \textbf{comfortable} }\\
Baseline & \textit{if      we      go      we      like    check   until the law that you know to do                                } & \textit{i       know   that is      much    more    \textbf{comes of one } }\\
Ours     & \textit{if      we      go      we      like    check   into    a       law if it does a job hotel                       } & \textit{i       know    that is      much    more  \textbf{  comfortable} }\\
\hline
\end{tabular}}
\end{table*}


\begin{table}[h]
\centering
\caption{Word Error Rate (WER) on the Switchboard dataset. None of our experiments used any lexicon information or external text data other than the training transcription. The models were trained on 300 hours of Switchboard data only.}
\label{tab:res}
\begin{tabular}{|r|r|r|}

\hline
Train ($\sim$ 300hrs)          & CH & SWB \\
Models                             & WER     & WER \\
\hline
\hline
\textit{\textbf{sentence-level end2end}} & & \\
Seq2Seq A2C \cite{zenkel2017comparison} & 40.6  & 28.1 \\
CTC A2C \cite{hannun2014deep} &  31.8 & 20.0 \\
CTC A2C \cite{zweig2017advances} & 32.1 & 19.8  \\
CTC A2W(Phone/external-LM init.) \cite{audhkhasi2017building} & 23.6  & 14.6 \\
\hline
\hline
\textit{\textbf{sentence-level end2end}} & & \\
Our baseline (CTC/Seq2Seq) & 34.4 & 19.0 \\
\textit{\textbf{dialog-context aware end2end}} & & \\
Our proposed model(a) & 34.1   & \textbf{18.2}        \\
Our proposed model(b) & \textbf{33.2}   & 18.6        \\
\hline
\end{tabular}
\end{table}

\begin{table}[h]
\centering
\caption{Substitution rate (Sub), Deletion rate (Del), and Insertion rate (Ins) for the baseline and our proposed model. }
\label{tab:wer}
\begin{tabular}{|r|r|r|r|r|r|}

\hline
Model              & Test & Sub  & Del & Ins & WER  \\
\hline
\hline
Baseline           & CH        & 23.9 & \textbf{5.8} & 4.7 & 34.4 \\
Proposed model(a) & CH        & 23.9 & 5.9 & 4.3 & 34.1 \\
Proposed model(b) & CH        & \textbf{22.8} & 6.3 & \textbf{4.1} & \textbf{33.2} \\
\hline
Baseline          & SWB       & 13.1 & 3.4 & 2.5 & 19.0   \\
Proposed model(a) & SWB       & \textbf{12.5} & 3.4 & \textbf{2.2} & \textbf{18.2} \\
Proposed model(b) & SWB       & 12.6 & 3.6 & 2.4 & 18.6 \\
\hline
\end{tabular}
\end{table}


\subsection{Training and decoding}
We used joint CTC/Attention end-to-end speech recognition architecture \cite{kim2017joint, watanabe2017hybrid} with ESPnet toolkit \cite{watanabe2018espnet}. We used a CNN-BLSTM encoder as suggested in \cite{zhang2017very, hori2017advances}. We followed the same six-layer CNN architecture as the prior study, except we used one input channel instead of three, since we did not use delta or delta delta features. Input speech features were downsampled to (1/4 x 1/4) along with the time-frequency axis. Then, the 4-layer BLSTM with 320 cells was followed by the CNN. We used a location-based attention mechanism \cite{chorowski2015attention}, where 10 centered convolution filters of width 100 were used to extract the convolutional features.

The decoder network of both our proposed models and the baseline models was a one-layer LSTM with 300 cells. Our dialog-context aware models additionally requires one-layer with 300 hidden units for incorporating the context vector with decoder states, and attention network with 2402-dimensional output layer to generate the context vector. We also built a character-level RNNLM (Char-RNNLM) on the the same Switchboard text dataset. The Char-RNNLM network was a two-layer LSTM with 650 cells, trained separately only on the training transcription. This network was used only for decoding. Note that we did not use any extra text data other than the training transcription.

The AdaDelta algorithm \cite{zeiler2012adadelta} with gradient clipping \cite{pascanu2013difficulty} was used for optimization. We used $\lambda = 0.5$ for joint CTC/Attention training. We bootstrap the training our proposed dialog-context aware end-to-end models from the baseline end-to-end models.  For decoding of the models, we used joint decoder which combines the output label scores from the AttentionDecoder, CTC, and Char-RNNLM \cite{hori2017advances} . 
%
%
The scaling factor of CTC, and RNNLM scores were $\alpha = 0.3$, and $\beta = 0.3$, respectively. We used a beam search algorithm similar to \cite{sutskever2014sequence} with the beam size 20 to reduce the computation cost. We adjusted the score by adding a length penalty, since the model has a small bias for shorter utterances. The final score $s(\bm{y}|\bm{x})$ is normalized with a length penalty $0.1$.
%

The models were implemented by using the Chainer deep learning library \cite{tokui2015chainer}, and ESPnet toolkit \cite{kim2017joint, watanabe2017hybrid, watanabe2018espnet}.

\vspace{0.5cm}

\section{RESULTS}
\label{sec:result}

We evaluated both the end-to-end speech recognition model which was built on sentence-level data (\textit{sentence-level end2end}) and our proposed dialog-context aware end-to-end speech recognition model which leveraged dialog-context information within and beyond the sentence (\textit{dialog-context aware end2end}).

Table \ref{tab:res} shows the WER of our baseline, proposed models, and several other published results those were only trained on 300 hours Switchboard training data. Note that CTC A2W(Phone/external-LM init.) \cite{audhkhasi2017building} was initialized from Phone CTC model and use external word embeddings.
As shown in Table \ref{tab:res}, we obtained a performance gain over our baseline \textit{sentence-level end2end} by using the dialog-context information. Our proposed model (a) performed best on SWB evaluation set showing 4.2\% relative improvement over our baseline. Our proposed model (b) performed best on CH evaluation set showing 3.4\% relative improvement over our baseline. 

Table \ref{tab:wer} shows the insertion, deletion, and substitution rates. We observed that the largest factor of WER improvement was from the substitution rates rather than deletion or insertion. Table \ref{tab:example} shows three example utterances with each previous sentence to show that our context information beyond the sentence-level improves word accuracy.

\section{CONCLUSION}
\label{sec:conclusion}
We proposed a dialog-context aware end-to-end speech recognition model which explicitly uses context information beyond sentence-level information. A key aspect of this model is that the whole system can be trained with dialog-level context information in an end-to-end manner. Our model was shown to outperform previous end-to-end models trained on sentence-level data. Moving forward, we plan to explore additional methods to represent the context information and evaluate performance improvements that can be obtained by addressing overfitting and data sparsity using larger conversational datasets, e.g., 2,000 hours of fisher dataset.

\section{ACKNOWLEDGEMENTS}
\label{sec:ack}

We gratefully acknowledge the support of NVIDIA Corporation with the donation of the Titan X Pascal GPU used for this research. This research was supported by a fellowship from the Center for Machine Learning and Health (CMLH) at Carnegie Mellon University. 


\bibliographystyle{IEEEbib}
\bibliography{strings,refs,mybib}

\begin{thebibliography}{10}

\bibitem{mikolov2010recurrent}
Tom{\'a}{\v{s}} Mikolov, Martin Karafi{\'a}t, Luk{\'a}{\v{s}} Burget, Jan
  {\v{C}}ernock{\`y}, and Sanjeev Khudanpur,
\newblock ``Recurrent neural network based language model,''
\newblock in {\em Eleventh Annual Conference of the International Speech
  Communication Association}, 2010.

\bibitem{mikolov2012context}
Tomas Mikolov and Geoffrey Zweig,
\newblock ``Context dependent recurrent neural network language model.,''
\newblock {\em SLT}, vol. 12, pp. 234--239, 2012.

\bibitem{wang2015larger}
Tian Wang and Kyunghyun Cho,
\newblock ``Larger-context language modelling,''
\newblock {\em arXiv preprint arXiv:1511.03729}, 2015.

\bibitem{ji2015document}
Yangfeng Ji, Trevor Cohn, Lingpeng Kong, Chris Dyer, and Jacob Eisenstein,
\newblock ``Document context language models,''
\newblock {\em arXiv preprint arXiv:1511.03962}, 2015.

\bibitem{liu2017dialog}
Bing Liu and Ian Lane,
\newblock ``Dialog context language modeling with recurrent neural networks,''
\newblock in {\em Acoustics, Speech and Signal Processing (ICASSP), 2017 IEEE
  International Conference on}. IEEE, 2017, pp. 5715--5719.

\bibitem{blei2003latent}
David~M Blei, Andrew~Y Ng, and Michael~I Jordan,
\newblock ``Latent dirichlet allocation,''
\newblock {\em Journal of machine Learning research}, vol. 3, no. Jan, pp.
  993--1022, 2003.

\bibitem{xiong2017microsoft}
Wayne Xiong, Jasha Droppo, Xuedong Huang, Frank Seide, Mike Seltzer, Andreas
  Stolcke, Dong Yu, and Geoffrey Zweig,
\newblock ``The microsoft 2016 conversational speech recognition system,''
\newblock in {\em Acoustics, Speech and Signal Processing (ICASSP), 2017 IEEE
  International Conference on}. IEEE, 2017, pp. 5255--5259.

\bibitem{graves2006connectionist}
Alex Graves, Santiago Fern{\'a}ndez, Faustino Gomez, and J{\"u}rgen
  Schmidhuber,
\newblock ``Connectionist temporal classification: labelling unsegmented
  sequence data with recurrent neural networks,''
\newblock in {\em Proceedings of the 23rd international conference on Machine
  learning}. ACM, 2006, pp. 369--376.

\bibitem{graves2014towards}
Alex Graves and Navdeep Jaitly,
\newblock ``Towards end-to-end speech recognition with recurrent neural
  networks,''
\newblock in {\em Proceedings of the 31st International Conference on Machine
  Learning (ICML-14)}, 2014, pp. 1764--1772.

\bibitem{hannun2014deep}
Awni Hannun, Carl Case, Jared Casper, Bryan Catanzaro, Greg Diamos, Erich
  Elsen, Ryan Prenger, Sanjeev Satheesh, Shubho Sengupta, Adam Coates, et~al.,
\newblock ``Deep speech: Scaling up end-to-end speech recognition,''
\newblock {\em arXiv preprint arXiv:1412.5567}, 2014.

\bibitem{miao2015eesen}
Yajie Miao, Mohammad Gowayyed, and Florian Metze,
\newblock ``{EESEN}: End-to-end speech recognition using deep {RNN} models and
  {WFST}-based decoding,''
\newblock in {\em 2015 IEEE Workshop on Automatic Speech Recognition and
  Understanding (ASRU)}. IEEE, 2015, pp. 167--174.

\bibitem{zweig2017advances}
Geoffrey Zweig, Chengzhu Yu, Jasha Droppo, and Andreas Stolcke,
\newblock ``Advances in all-neural speech recognition,''
\newblock in {\em Acoustics, Speech and Signal Processing (ICASSP), 2017 IEEE
  International Conference on}. IEEE, 2017, pp. 4805--4809.

\bibitem{bahdanau2014neural}
Dzmitry Bahdanau, Kyunghyun Cho, and Yoshua Bengio,
\newblock ``Neural machine translation by jointly learning to align and
  translate,''
\newblock {\em arXiv preprint arXiv:1409.0473}, 2014.

\bibitem{chorowski2014end}
Jan Chorowski, Dzmitry Bahdanau, Kyunghyun Cho, and Yoshua Bengio,
\newblock ``End-to-end continuous speech recognition using attention-based
  recurrent {NN}: First results,''
\newblock {\em arXiv preprint arXiv:1412.1602}, 2014.

\bibitem{chorowski2015attention}
Jan~K Chorowski, Dzmitry Bahdanau, Dmitriy Serdyuk, Kyunghyun Cho, and Yoshua
  Bengio,
\newblock ``Attention-based models for speech recognition,''
\newblock in {\em Advances in Neural Information Processing Systems}, 2015, pp.
  577--585.

\bibitem{chan2015listen}
William Chan, Navdeep Jaitly, Quoc~V Le, and Oriol Vinyals,
\newblock ``Listen, attend and spell,''
\newblock {\em arXiv preprint arXiv:1508.01211}, 2015.

\bibitem{kim2017joint}
Suyoun Kim, Takaaki Hori, and Shinji Watanabe,
\newblock ``Joint ctc-attention based end-to-end speech recognition using
  multi-task learning,''
\newblock in {\em Acoustics, Speech and Signal Processing (ICASSP), 2017 IEEE
  International Conference on}. IEEE, 2017, pp. 4835--4839.

\bibitem{watanabe2017hybrid}
Shinji Watanabe, Takaaki Hori, Suyoun Kim, John~R Hershey, and Tomoki Hayashi,
\newblock ``Hybrid ctc/attention architecture for end-to-end speech
  recognition,''
\newblock {\em IEEE Journal of Selected Topics in Signal Processing}, vol. 11,
  no. 8, pp. 1240--1253, 2017.

\bibitem{swbd}
John~J Godfrey, Edward~C Holliman, and Jane McDaniel,
\newblock ``Switchboard: Telephone speech corpus for research and
  development,''
\newblock {\em Acoustics, Speech, and Signal Processing, 1992. ICASSP-92., 1992
  IEEE International Conference on}, 1992.

\bibitem{godfrey1992switchboard}
John~J Godfrey, Edward~C Holliman, and Jane McDaniel,
\newblock ``Switchboard: Telephone speech corpus for research and
  development,''
\newblock in {\em Acoustics, Speech, and Signal Processing, 1992. ICASSP-92.,
  1992 IEEE International Conference on}. IEEE, 1992, vol.~1, pp. 517--520.

\bibitem{bahdanau2015end}
Dzmitry Bahdanau, Jan Chorowski, Dmitriy Serdyuk, Philemon Brakel, and Yoshua
  Bengio,
\newblock ``End-to-end attention-based large vocabulary speech recognition,''
\newblock {\em arXiv preprint arXiv:1508.04395}, 2015.

\bibitem{zenkel2017comparison}
Thomas Zenkel, Ramon Sanabria, Florian Metze, Jan Niehues, Matthias Sperber,
  Sebastian St{\"u}ker, and Alex Waibel,
\newblock ``Comparison of decoding strategies for ctc acoustic models,''
\newblock {\em arXiv preprint arXiv:1708.04469}, 2017.

\bibitem{audhkhasi2017building}
Kartik Audhkhasi, Brian Kingsbury, Bhuvana Ramabhadran, George Saon, and
  Michael Picheny,
\newblock ``Building competitive direct acoustics-to-word models for english
  conversational speech recognition,''
\newblock {\em arXiv preprint arXiv:1712.03133}, 2017.

\bibitem{watanabe2018espnet}
Shinji Watanabe, Takaaki Hori, Shigeki Karita, Tomoki Hayashi, Jiro Nishitoba,
  Yuya Unno, Nelson Enrique~Yalta Soplin, Jahn Heymann, Matthew Wiesner, Nanxin
  Chen, et~al.,
\newblock ``Espnet: End-to-end speech processing toolkit,''
\newblock {\em arXiv preprint arXiv:1804.00015}, 2018.

\bibitem{zhang2017very}
Yu~Zhang, William Chan, and Navdeep Jaitly,
\newblock ``Very deep convolutional networks for end-to-end speech
  recognition,''
\newblock in {\em Acoustics, Speech and Signal Processing (ICASSP), 2017 IEEE
  International Conference on}. IEEE, 2017, pp. 4845--4849.

\bibitem{hori2017advances}
Takaaki Hori, Shinji Watanabe, Yu~Zhang, and William Chan,
\newblock ``Advances in joint ctc-attention based end-to-end speech recognition
  with a deep cnn encoder and rnn-lm,''
\newblock {\em arXiv preprint arXiv:1706.02737}, 2017.

\bibitem{zeiler2012adadelta}
Matthew~D Zeiler,
\newblock ``Adadelta: an adaptive learning rate method,''
\newblock {\em arXiv preprint arXiv:1212.5701}, 2012.

\bibitem{pascanu2013difficulty}
Razvan Pascanu, Tomas Mikolov, and Yoshua Bengio,
\newblock ``On the difficulty of training recurrent neural networks,''
\newblock in {\em International Conference on Machine Learning}, 2013, pp.
  1310--1318.

\bibitem{sutskever2014sequence}
Ilya Sutskever, Oriol Vinyals, and Quoc~VV Le,
\newblock ``Sequence to sequence learning with neural networks,''
\newblock in {\em Advances in neural information processing systems}, 2014, pp.
  3104--3112.

\bibitem{tokui2015chainer}
Seiya Tokui, Kenta Oono, Shohei Hido, and Justin Clayton,
\newblock ``Chainer: a next-generation open source framework for deep
  learning,''
\newblock in {\em Proceedings of Workshop on Machine Learning Systems
  (LearningSys) in The Twenty-ninth Annual Conference on Neural Information
  Processing Systems (NIPS)}, 2015.

\end{thebibliography}

\end{document}